\documentclass{article}
\usepackage{amsmath,graphicx,amssymb}
\usepackage{multirow}
\usepackage{pifont}
\usepackage{cite}

\usepackage[preprint]{spconf}
\copyrightnotice{\copyright\ IEEE 2024}
\toappear{To appear in {\it Proc.\ ICIP2024, October 27-30, 2024, Abu Dhabi, United Arab Emirates}}

\usepackage{ifthen}
\usepackage{atbegshi}
\newcounter{pagecounter}
\AtBeginShipout{%
  \stepcounter{pagecounter}%
  \ifthenelse{\value{pagecounter}=1}{%
    \AtBeginShipoutUpperLeft{%
      \put(\dimexpr\paperwidth-20.3cm\relax,-0.8cm){%
        \parbox[t]{19cm}{%
          \copyright\ 2024 IEEE. Personal use of this material is permitted. 
          Permission from IEEE must be obtained for all other uses, in any current or future media, 
          including reprinting/republishing this material for advertising or promotional purposes, 
          creating new collective works, for resale or redistribution to servers or lists, 
          or reuse of any copyrighted component of this work in other works.%
        }%
      }%
    }%
  }{}%
}

\newcommand{\cmark}{\ding{51}}%
\newcommand{\xmark}{\ding{55}}%
\title{YOLO-FEDER FusionNet: A Novel Deep Learning Architecture for Drone Detection}
\name{Tamara R. Lenhard$^{1,2,}$\sthanks{Corresponding Author, E-mail: tamara.lenhard@dlr.de}, Andreas Weinmann$^{2}$, Stefan Jäger$^{1}$ and Tobias Koch$^{1}$}
  
\address{$^{1}$ {\normalsize Institute for the Protection of Terrestrial Infrastructures, German Aerospace Center (DLR), Sankt Augustin, Germany}\\
$^{2}$ {\normalsize Working Group Algorithms for Computer Vision, Imaging and Data Analysis, University of Applied Sciences Darmstadt, Darmstadt,}\\ {\normalsize Germany}}

\begin{document}
\ninept

\maketitle

% ABSTRACT
\begin{abstract}
Predominant methods for image-based drone detection frequently rely on employing generic object detection algorithms like YOLOv5. While proficient in identifying drones against homogeneous backgrounds, these algorithms often struggle in complex, highly textured environments. In such scenarios, drones seamlessly integrate into the background, creating camouflage effects that adversely affect the detection quality. To address this issue, we introduce a novel deep learning architecture called YOLO-FEDER FusionNet. Unlike conventional approaches, YOLO-FEDER FusionNet combines generic object detection methods with the specialized strength of camouflage object detection techniques to enhance drone detection capabilities. Comprehensive evaluations of YOLO-FEDER FusionNet show the efficiency of the proposed model and demonstrate substantial improvements in both reducing missed detections and false alarms.
\end{abstract}

% KEYWORDS

% INTRODUCTION
\section{Introduction}
\label{sec:intro}
Robust drone detection systems play a vital role in enhancing security systems, protecting privacy and ensuring regulatory compliance~\cite{Chiper:2022}. Leveraging advanced computer vision techniques, image-based drone detection establishes a proactive mechanism to analyze visual data, facilitating early threat detection, and enabling effective mitigation measures. The widespread adoption of image-based detection techniques is primarily driven by the cost-effectiveness of camera sensors, their broad availability, and their seamless integration into established security systems~\cite{Elsayed:2021}. 

In the field of drone detection, the processing of acquired image data typically relies on the application of generic object detection models (e.g., YOLOv5~\cite{Ultralytics} in various network configurations). These models enjoy widespread popularity due to their adeptness in balancing real-time processing speed and precision. Furthermore, generic object detection models exhibit notable effectiveness in detecting drones against homogeneous backgrounds (e.g., clear blue sky), or in scenarios where drones distinctly contrast with their surroundings~\cite{Seidaliyeva:2020}. However, their performance tends to decline considerably in scenarios where drones operate against complex and highly textured backgrounds~\cite{Seidaliyeva:2020,Dieter:2023}. Our previous investigations specifically emphasized the substantial challenge inherent in detecting drones amidst or in close proximity to trees. The heterogeneous background composition, combined with difficult lighting conditions and the similarity between tree branches and drone rotor arms, facilitates a seamless integration of drones into their surroundings. The resulting camouflage effect severely impedes the capacity of generic object detection systems to accurately identify and delineate the boundaries of drones, undermining the overall detection quality~\cite{Dieter:2023}. The phenomenon of camouflage effects extends beyond drone detection. For instance, it constitutes a considerable challenge in accurately detecting animals within their natural habitats, fostering the development of diverse camouflage object detection (COD) techniques~\cite{Fan:2020}. 

\begin{figure}
\centering
\begin{minipage}[b]{1.0\linewidth}
  \includegraphics[width=\textwidth, trim={0.1cm 16.2cm 0.1cm 0cm}, clip]{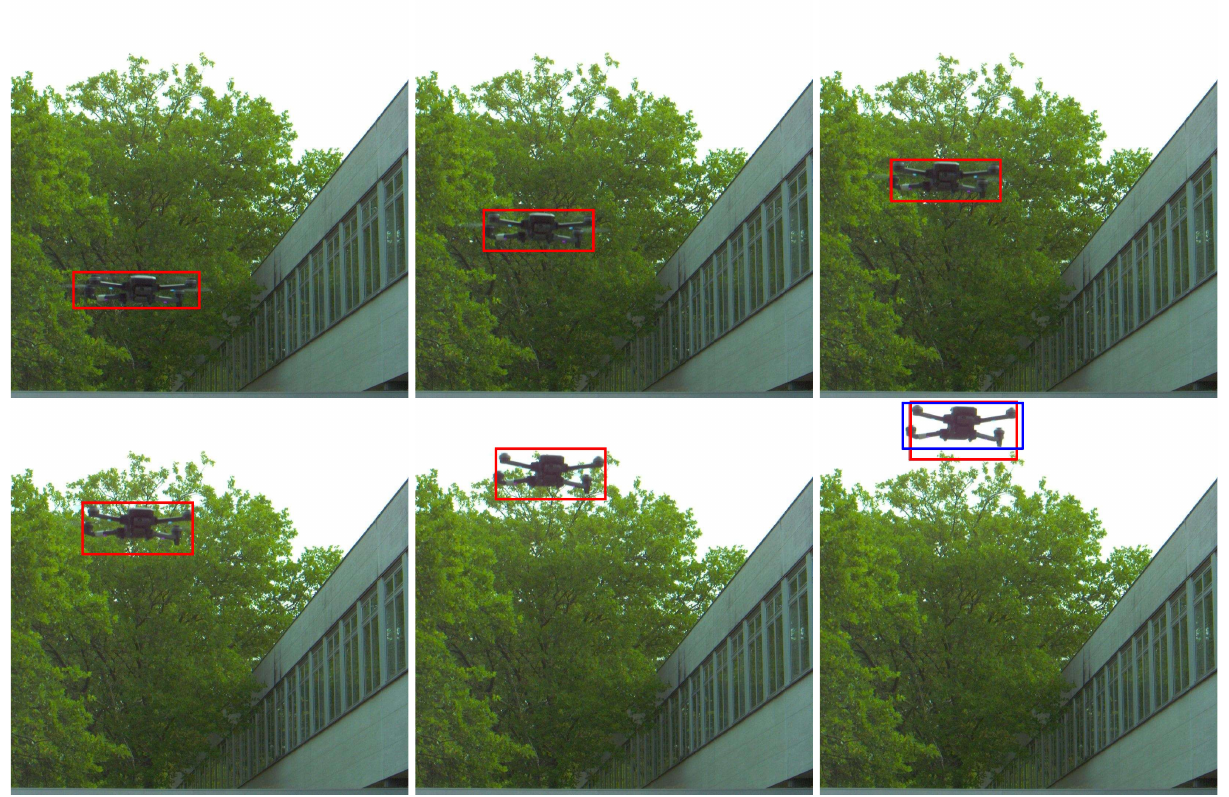}
  \caption{\label{fig:Comparison} Visual comparison between YOLO-FEDER FusionNet (red bounding boxes) and YOLOv5l (blue bounding box), showcasing every fifth image frame. YOLO-FEDER FusionNet consistently detects the drone across all six frames, whereas YOLOv5l only identifies it in the last one.}
\end{minipage}
\end{figure}

While COD techniques have demonstrated efficiency in animal detection, their direct transposition to drone detection is unexplored. Therefore, our study aims to assess the viability of leveraging insights from COD methods to enhance the reliability of generic drone detectors, especially in scenarios where they encounter limitations (cf. Figure~\ref{fig:Comparison}). We introduce YOLO-FEDER FusionNet, a novel deep learning (DL) architecture that combines the strengths of generic object detection with the specialized capabilities of COD. Furthermore, we provide an examination of YOLO-FEDER FusionNet across diverse real-world datasets. This also entails a comparative analysis against established drone detection techniques, offering a robust evaluation of the effectiveness and performance enhancements achieved by our approach. Additionally, we introduce a simple techniques for false negative mitigation in image sequences.

Given the limited availability of annotated real-world data for drone detection and the problem-specific nature of existing datasets, our method strategically incorporates synthetic data -- a prevalent practice to mitigate this scarcity~\cite{Barisic:2022,Marez:2020,Symeonidis:2022}. Nevertheless, there is still a discrepancy between synthetically generated and manually labeled real-world data. Consequently, our study also addresses the gap between simulated scenarios and real-world conditions, especially focusing on the inherent bias induced by manual labeling procedures.

The remainder of the paper is structured as follows: Section~\ref{sec:relatedwork} presents a review of the current state-of-the-art, followed by a detailed exposition of the proposed drone detection framework (Section~\ref{sec:framework}). Section~\ref{sec:expsetup} describes the experimental setup including datasets, evaluation metrics, and implementation details. The results are presented and discussed in Section~\ref{sec:results}. In Section~\ref{sec:conclusion}, we draw conclusions.

% RELATED WORK
\section{Related Work}
\label{sec:relatedwork}
In the following, we discuss the latest advancements in image-based drone detection, focusing on their effectiveness in challenging environments. Additionally, we address the principle idea of COD along with prevalent COD techniques. Furthermore, we explore essential research and concepts related to generating synthetic data, addressing the inherent discrepancies between simulated and real-world scenarios in the context of drone detection.

\paragraph*{Drone Detection.} Developing precise and reliable drone detection systems is a multidimensional challenge, accommodating various interpretations, sub-problems and strategic directions. Within the domain of image-based drone detection, considerable emphasis is directed towards the detection of small drones~\cite{Lv:2022,Liu:2021} and accurately discerning them from other aerial entities, such as birds~\cite{Lv:2022}. A notable deficiency persists in methodologies specifically tailored to the enhancement of drone detection amidst complex or highly textured backgrounds~\cite{Lv:2022,Dieter:2023}. Recent strategies addressing this challenge commonly entail the refinement of distinct modules within established object detection frameworks, such as YOLOv5~\cite{Lv:2022}. For instance, Lv et al.~\cite{Lv:2022} introduce SAG-YOLOv5s -- an adapted smaller version of YOLOv5s (denoting the small variant within the YOLOv5 series), specifically optimized for intricate environments. Their method integrates SimAM attention~\cite{Yang:2021} and Ghost modules~\cite{Han:2020} into YOLOv5s' bottleneck structure, to elevate drone target extraction and refine background suppression during feature analysis. Concurrently, alternative methodologies seek to simplify the complexity of natural environments by dividing the process into two distinct stages. Typically, this involves eliminating background elements~\cite{Seidaliyeva:2020} and extracting moving objects~\cite{Chen:2017}, followed by a classification. Furthermore, some methodologies contemplate camera-based drone detection either as a preliminary stage in a tracking procedure or an integral part of a multi-sensor system~\cite{Svanstroem:2022}. This strategic consideration aims to offset potential shortcomings of a purely camera-based system, augmenting its robustness especially in complex environments. However, the challenge of camouflage effects induced by natural elements like trees remains unaddressed in current drone detection strategies.

\paragraph*{Camouflage Object Detection.} The development of specialized methodologies dedicated to the detection of camouflage objects is an emerging field of research. Camouflage object detection (COD) embodies a class-independent detection task~\cite{Fan:2020}, particularly prevalent in the domain of animal detection. Its objective is the precise identification of objects that closely replicate the inherent characteristics of their surrounding, minimizing their visual contrast and distinctiveness. The majority of COD techniques address the challenges posed by intrinsic similarity and edge disruption through emulation of the human visual system~\cite{Fan:2020,Jia:2022}. Only a small selection of approaches seeks to compensate for perception limitations by disassembling the camouflage scenario and emphasizing subtle distinguishing features~\cite{He:2023}. A promising model belonging to the second algorithmic category is the feature decomposition and edge reconstruction (FEDER) model by He et al.~\cite{He:2023}.

\begin{figure*}[htb!]
\centering
\begin{minipage}[b]{0.95\linewidth}
  \includegraphics[width=\textwidth, trim={0.3cm 20.35cm 0.3cm 0cm}, clip]{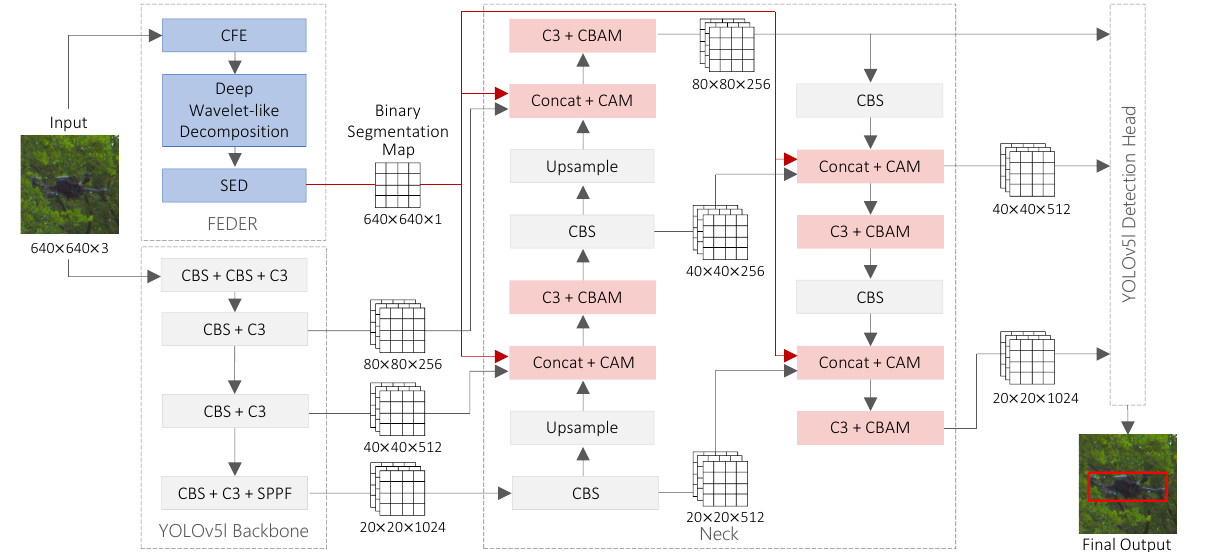}
  \caption{\label{fig:YoloFeder} Overview of YOLO-FEDER FusionNet. Key components fusing and processing information from both backbones are highlighted in red. Layers are abbreviated as follows: CFE (camouflage feature encoder), CAM (channel attention module), CBS (convolution + batch normalization + SiLU activation), CBAM (convolutional block attention module), C3 (CSP bottleneck with three convolutional layers), SED (segmentation-oriented edge-assisted decoder), SPPF (spatial pyramid pooling fusion). The visualization is inspired by the illustration of the YOLOv5l architecture in~\cite{Ultralytics}.}
\end{minipage}
\end{figure*}

\paragraph*{Simulation-Reality Gap.} Leveraging synthetic data is a popular approach for training DL models in drone detection~\cite{Barisic:2022,Marez:2020,Symeonidis:2022,Dieter:2023} and other application domains~\cite{Pham:2022} due to the high expenses tied to acquiring real-world data. Techniques like domain randomization~\cite{Marez:2020} or game engine-based simulations~\cite{Dieter:2022} facilitate the generation of extensive, domain-specific datasets. These methods demonstrate cost-effectiveness through automated labeling processes, ensuring precise annotations, unlike manual labeling techniques. Furthermore, they enable the circumvention of real-world constraints (e.g., privacy regulations), fostering dataset diversification. However, transferring detection models exclusively trained on synthetic data to real-world applications frequently leads to performance degradation, attributed to the simulation-reality gap. The gap's severity is closely linked to the quality of both synthetic and real-world data, and is typically assessed through diverse quality measures such as mAP across various intersection over union (IoU) thresholds~\cite{Barisic:2022,Marez:2020}. Two primary strategies for narrowing this gap include fine-tuning with real-world data~\cite{Marez:2020} and mixed-data training~\cite{Chen:2017}.

% FRAMEWORK
\section{Framework}
\label{sec:framework} 
Given the complexity inherent to drone detection, the proposed YOLO-FEDER FusionNet strategically combines the benefits of generic object detection with the specific strengths of COD algorithms. The model relies on two essential components for feature extraction: the well-established YOLOv5l backbone architecture~\cite{Ultralytics} and the specialized camouflage object detector FEDER~\cite{He:2023}. YOLOv5l refers to the larger model configuration within the YOLOv5 series. As both algorithms yield complementary results, the YOLOv5l backbone and the FEDER algorithm operate as an ensemble system to extract significant features. This involves parallel processing an RGB image $\mathbf{X}\in \mathbb{R}^{W\times H\times 3}$ with $W=H$, by both components (cf. Figure~\ref{fig:YoloFeder}). The information from both components is fused at feature level within the network's neck, whose architectural design is inspired by YOLOv5l~\cite{Ultralytics}. The feature maps issued by the neck are processed within the network's head to generate predictions for objects across three distinct sizes. Detailed descriptions of all network components are presented in the following sections. 

\paragraph*{YOLOv5l Backbone.} The employed YOLOv5l backbone~\cite{Ultralytics} relies on CSPDarkNet53, which incorporates DarkNet-53~\cite{Redmon:2018} in conjunction with an advanced CSPNet strategy~\cite{Wang:2020}. The foundational architecture features a sequential arrangement of multiple CBS modules (consisting of convolutional, batch normalization, and SiLU activation layers) and C3 modules (comprising a CSP bottleneck with three convolutional layers). A spatial pyramid pooling fusion (SPPF) module~\cite{He:2014} completes the backbone structure (see Figure~\ref{fig:YoloFeder}, bottom left).

\paragraph*{FEDER Backbone.} The feature decomposition and edge reconstruction (FEDER) model introduced by He et al.~\cite{He:2023} consists of three main components: a camouflaged feature encoder (CFE), a deep wavelet-like decomposition (DWD) module, and a segmentation-oriented edge-assisted decoder (SED) (see Figure~\ref{fig:YoloFeder}, top left). The CFE leverages a Res2Net50~\cite{Gao:2019}, in combination with R-Net~\cite{Fan:2020}, to generate a series of feature maps given an input image $\mathbf{X}\in \mathbb{R}^{W\times H\times 3}$, with $W=H$. These feature maps serve as inputs for an efficient atrous spatial pyramid pooling (e-ASPP) module~\cite{Chen:2016} and the DWD. As discriminative attributes in COD primarily reside in high-frequency (HF) and low-frequency (LF) components~\cite{Stevens:2009} -- such as texture and edges (HF), as well as color and illumination (LF) -- the feature maps within the DWD module are partitioned into distinct HF and LF parts. The partitioning process involves employing learnable HF and LF filters, coupled with adaptive wavelet distillation~\cite{Ha:2021} for updating the coefficients. Furthermore, the DWD leverages HF and LF attention modules, alongside guidance-based feature aggregation, to systematically extract discriminative information from the decomposed features and fuse this information in a meaningful way. The features derived from the DWD and the e-ASPP are decoded via the SED. Within the SED, a reversible re-calibration segmentation (RRS) module and an ordinary differential equation (ODE)-inspired edge reconstruction (OER) module are employed for sophisticated feature processing and auxiliary edge reconstruction. The final output generated by the FEDER backbone comprises a binary segmentation map $\mathbf{O}_S\in \mathbb{R}^{W\times H\times 1}$ and an edge prediction map $\mathbf{O}_E\in \mathbb{R}^{W\times H\times 1}$ with $W=H$. In YOLO-FEDER FusionNet, only the segmentation map is further processed. For more details on FEDER, refer to~\cite{He:2023}.

\paragraph*{Neck.} The neck of YOLO-FEDER FusionNet is specifically designed to unify information from both backbones across diverse layers (cf. Figure~\ref{fig:YoloFeder}). Drawing inspiration from the foundational architecture of YOLOv5l~\cite{Ultralytics}, the neck's architectural design features CBS, C3, and upsampling modules (akin to YOLOv5l). In addition, modified concatenation layers are incorporated to facilitate the integration of outputs from the FEDER backbone (cf. Figure~\ref{fig:YoloFeder}, red connections), effectively complementing information gathered from preceding layers (cf. Figure~\ref{fig:YoloFeder}, gray connections). Furthermore, attention mechanisms are strategically embedded at multiple positions within the network's neck (cf. Figure~\ref{fig:YoloFeder}, red components) to enable the prioritization of significant features. Attention mechanisms frequently only concentrate on either spatial or channel-related feature relationships (cf.~\cite{Guo:2021}). A widely adopted attention mechanism, combining spatial and channel-wise attention, is the convolutional block attention module (CBAM) introduced by Woo et al.~\cite{Woo:2018}.

\begin{figure}
\centering
\begin{minipage}[b]{1\linewidth}
  \includegraphics[width=\textwidth, trim={2.3cm 24.4cm 2.1cm 0cm}, clip]{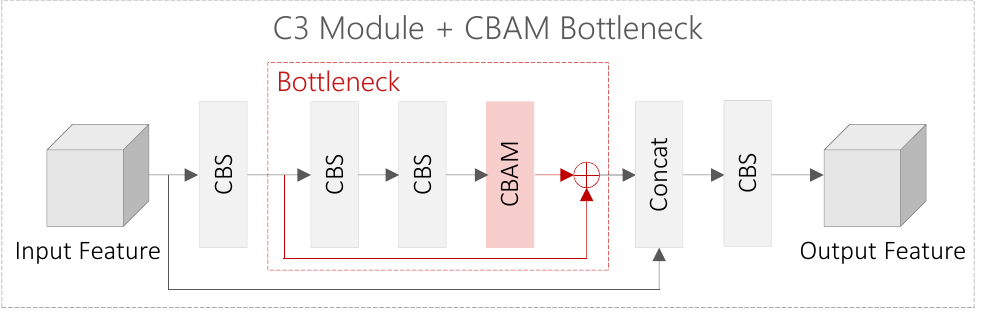}
  \caption{\label{fig:C3plusCBAM} Integration of CBAM into the bottleneck of the C3 module, located by default in the neck and head of YOLOv5l. Modified parts are highlighted in red.}
\end{minipage}
\end{figure}

Inspired by Woo et al.'s Res50 + CBAM model~\cite{Woo:2018}, where CBAM was embedded within the residual blocks, we integrated this module into our proposed network architecture in a similar way. Precisely, it is located within the CSP bottleneck of the C3 module (cf. Figure~\ref{fig:YoloFeder}, C3 + CBAM). Considering an intermediate feature map $\mathbf{F}\in \mathbb{R}^{W\times H\times C}$ derived from a preceding CBS module (cf. Figure~\ref{fig:C3plusCBAM}), the overall attention process initiated by CBAM can be described as follows: \vspace{-0.1cm}
\begin{center} $\mathbf{F}' = \mathbf{M}_S(\mathbf{M}_C(\mathbf{F})\otimes \mathbf{F}) \otimes (\mathbf{M}_C(\mathbf{F})\otimes \mathbf{F})$ .\end{center}

\noindent
Here, $\mathbf{M}_C(\mathbf{F})\in \mathbb{R}^{1\times 1\times C}$ represents the one-dimensional channel attention map, $\mathbf{M}_S(\mathbf{F})\in \mathbb{R}^{W\times H\times 1}$ the two-dimensional spatial attention map, $\otimes$ denotes the element-wise multiplication, and $\mathbf{F}'\in \mathbb{R}^{W\times H\times C}$ signifies the refined feature map. Note that during multiplication, $\mathbf{M}_C(\mathbf{F})$ is replicated along the spatial dimensions, while $\mathbf{M}_S(\mathbf{F})$ is duplicated along the channel dimension~\cite{Woo:2018}. The integration of CBAM aims to direct the model's attention towards relevant areas and optimize its focus. Additionally, we implemented a channel-wise attention mechanism (akin to the one featured in CBAM, cf.~\cite{Woo:2018} for details) following the concatenation of information across diverse layers. For instance, when linking an intermediate feature map $\mathbf{F}\in \mathbb{R}^{W\times H\times C}$ obtained by the YOLOv5l backbone with the binary segmentation map $\mathbf{O}_S\in \mathbb{R}^{W\times H\times 1}$ derived from FEDER, the instantiated attention mechanism can be described as follows:\vspace{-0.1cm}

\begin{center}
$\mathbf{F}' = \mathbf{M}_C(\mathrm{Concat}(\mathbf{F}, \mathbf{O}_S ))\otimes \mathrm{Concat}(\mathbf{F}, \mathbf{O}_S )$
\end{center}

\noindent
where $\mathbf{M}_C(\mathbf{F})$ is once again replicated along the spatial dimensions. This mechanism aims to account for inter-dependencies and relationships among different channels within a feature map. Therefore, it is particular beneficial when consolidating data from multiple sources via concatenation, facilitating the selection and prioritization of the most relevant details from each source.

\paragraph*{Head.} The head of the proposed network replicates the design of the standard YOLOv5l head. Its primary function involves predicting objects across three distinct sizes (small, medium, and large). 

% EXPERIMENTAL SETUP
\section{Experimental Setup}
\label{sec:expsetup}
To assess the proposed framework, we employ the following experimental setup, incorporating diverse datasets and evaluation metrics.
\subsection{Datasets}
\label{subsec:Data} 
Considering the scarcity of accessible data in the context of drone detection, we utilize self-captured real-world data sourced from a potential application site for evaluation. Concurrently, we leverage synthetically generated data, derived from physically-realistic simulations, to effectively train the proposed detection model. Table~\ref{tab:data} provides an overview of the datasets employed in this study, with details discussed below. Both real and synthetically generated data have already been included in our prior work~\cite{Dieter:2023}. 

\paragraph*{Real-World Data.} To acquire real-world data, we employ a fixed Basler acA200-165c camera system firmly stationed on the ground. The system is equipped with dual lenses (25mm and 8mm), enabling the capture of two distinct field of views from each vantage point. The selected recording environment replicates structural and environmental characteristics of a potential installation site for a drone detection system in an urban surveillance setting (cf.~\cite{Dieter:2023} for more details). The original RGB images are recorded at a resolution of 2040$\times$1086 pixels, leading to two distinct datasets R1 and R2 (cf. Table~\ref{tab:data}). While the background composition in dataset R1 is characterized by an increased prevalence of building structures, highly textured objects -- more precisely trees -- form a substantial part of the image backgrounds in dataset R2. Thus, R2 exhibits a greater level of complexity in comparison to R1. Given the model's constraint necessitating square images, a coarse cropping strategy is deployed, contingent upon the precise localization of the drone object within the image frame. Subsequently, a random cropping technique is applied using distinct dimensions: 640$\times$640 (YOLOv5l's default input size) and 1080$\times$1080. This procedure yields two different versions of each dataset: one set featuring images at a resolution of 640$\times$640 pixels, and another set comprising images of size 1080$\times$1080 pixels (to amplify the informational content). Employing this systematic approach ensures the preservation of crucial information while concurrently enhancing dataset diversity. Alongside drone imagery, all datasets also include approximately 7-8~\% of background images. 

\paragraph*{Synthetic Data.} To generate synthetic training data, we employ the game engine-based data generation pipeline detailed in~\cite{Dieter:2022}. The pipeline harnesses the functionalities of the Unreal Engine version 4.27~\cite{UnrealEngine} and Microsoft AirSim~\cite{Airsim}, enabling the efficient extraction of automatically labeled RGB images. Leveraging the Urban City environment~\cite{UrbanCity}, we aim to emulate essential attributes of the application scenario defined by R1 and R2. Data collection is performed from five unique camera perspectives, employing three distinct drone models (see~\cite{Dieter:2023} for more details). Aligned with the characteristics of R1 and R2, synthetic RGB images are initially captured at a resolution of 2040$\times$1080 pixels (leading to dataset S1, cf. Table~\ref{tab:data}). Subsequently, a cropping procedure is applied (in analogy to R1 and R2) to achieve a final resolution of 640$\times$640 pixels. Dataset S1 also includes a small share of background images (7-8~\%).

\begin{table}[t!]
\caption{Overview of training and validation datasets.\hspace{1.7cm}}
\centering
  \label{tab:data}
  \footnotesize
  \begin{tabular}{cccccccc}
  \hline\noalign{\smallskip}
    Dataset & Type & \multicolumn{3}{c}{Image Count} & \multicolumn{2}{c}{Camera} \\
    & & Train & Val & Test & Pos. & Focal Length\\\noalign{\smallskip}\hline\noalign{\smallskip}
    R1 & real & 7,524 & 2,508 & 2,508 & 2 & 8 mm\\\noalign{\smallskip}\hline\noalign{\smallskip}
    R2 & real & 3,834 & 1,279 & 1,278 & 2 & 25 mm\\\noalign{\smallskip}\hline\noalign{\smallskip}
    S1 & synth. & 10,446 & 3,483 & 3,483 & 5 & -- \\\hline
  \end{tabular}
\end{table}

\subsection{Evaluation Metrics} 
Ensuring security against unauthorized drone intrusion necessitates precise early-stage detection. Thus, an exceptionally low false negative rate (FNR) is essential for a reliable detection system. However, in scenarios involving continuous data streams (e.g., commonly encountered in surveillance settings), detecting a drone in every frame of the sequence is not imperative. Extrapolating from adjacent frames enables (to a certain extent) partial inference of missed detections. Considering drone detection as an integral component of a comprehensive security framework, the reduction of false positives is also crucial for system credibility. This is akin to reducing the false discovery rate (FDR). Complementing the evaluation via FNR and FDR, we include the mean average precision (mAP) at an intersection over union (IoU) threshold of 0.5, due to its widespread adoption as key performance indicator for object detection models. As the requirement for precise bounding box localization can be alleviated in our application context and manually generated annotations exhibit notable variance in quality, we also consider mAP values at an IoU threshold of 0.25.

\subsection{Implementation Details} 
YOLO-FEDER FusionNet is implemented in PyTorch, leveraging the foundation of the original YOLOv5 framework provided by~\cite{Ultralytics}. It incorporates a YOLOv5l backbone pre-trained on the COCO benchmark dataset~\cite{Lin:2014}, as well as a FEDER network initialized with COD10K~\cite{Fan:2020} weights, both remaining frozen during training. The model's neck and head are optimized using stochastic gradient descent (SGD) with an initial learning rate of 0.01, a momentum of 0.937, and a weight decay of 0.005. In the training phase, we maintain a batch size of 32. We assume square input images that are uniformly resized to 640$\times$640 for training and inference. We deliberately avoid letter-boxing or random resizing to handle rectangular images. Additionally, we train two standard YOLOv5l models~\cite{Ultralytics} for comparative analysis, using the same hyper-parameter configuration as for YOLO-FEDER FusionNet. The first model is trained on dataset S1 in its original, un-cropped version, as in our prior work~\cite{Dieter:2023}. The second model (YOLOv5l SQ) is trained on the cropped version of S1 (cf. Section~\ref{subsec:Data}). In both scenarios, no layers are frozen during the training process. All experiments are conducted on a single NVIDIA Quadro RTX-8000 GPU.

% RESULTS
\section{Results}
\label{sec:results}
In this section, we present the evaluation results of our proposed framework, providing an examination of its performance on real-world data, the mitigation of labeling biases through a post-processing strategy, and the assessment of the framework's effectiveness in an alarm scenario.

\begin{table}[t!]
\centering
\caption{Evaluation and comparison of YOLOv5l, YOLOv5l SQ, and YOLO-FEDER FusionNet on dataset R1.}
  \label{tab:resultsR1}
  \footnotesize
  \begin{tabular}{lccccc}
  \hline\noalign{\smallskip}
    Model & Img Size & \multicolumn{2}{c}{mAP} & FNR & FDR \\
    & & @0.25 & @0.5 \\\noalign{\smallskip}\hline\noalign{\smallskip}
    YOLOv5l & 2040$\times$1086 & 0.572 & 0.551 & 0.463 & 0.500\\\noalign{\smallskip}\hline\noalign{\smallskip}
    YOLOv5l & \multirow{3}*{640$\times$640} & 0.433 & 0.401 & 0.601 & 0.311\\
    YOLOv5l SQ & & 0.209 & 0.191 & 0.913 & 0.033\\
    FusionNet (Ours) & & 0.729 & 0.669 &0.372 & 0.114\\\noalign{\smallskip}\hline\noalign{\smallskip}
    
    YOLOv5l & \multirow{3}*{1080$\times$1080} & 0.568 & 0.548 & 0.457 & 0.261\\
    YOLOv5l SQ & & 0.454 & 0.380 & 0.932 & 0.078\\
    FusionNet (Ours) & & 0.708 & 0.636 & 0.449 & 0.066\\
    \hline
  \end{tabular}
\end{table}

\begin{table}[t!]
\centering
\caption{Evaluation and comparison of YOLOv5l, YOLOv5l SQ, and YOLO-FEDER FusionNet on dataset R2.}
  \label{tab:resultsR2}
  \footnotesize
  \begin{tabular}{lccccc}
  \hline\noalign{\smallskip}
    Model & Img Size & \multicolumn{2}{c}{mAP} & FNR & FDR \\
    & & @0.25 & @0.5 \\\noalign{\smallskip}\hline\noalign{\smallskip}
    YOLOv5l & 2040$\times$1086 & 0.571 & 0.432 & 0.745 & 0.290\\\noalign{\smallskip}\hline\noalign{\smallskip}
    YOLOv5l & \multirow{3}*{640$\times$640} & 0.102 & 0.047 & 0.858 & 0.638\\
    YOLOv5l SQ & & 0.252 & 0.078 & 0.955 & 0.010\\
    FusionNet (Ours) & & 0.685 & 0.270 & 0.473 & 0.029\\\noalign{\smallskip}\hline\noalign{\smallskip}
    
    YOLOv5l & \multirow{3}*{1080$\times$1080} & 0.396 & 0.196 & 0.698 & 0.249\\
    YOLOv5l SQ & & 0.343 & 0.101 & 0.986 & 0.058\\
    FusionNet (Ours) & & 0.816 & 0.423 & 0.335 & 0.007\\
    \hline
  \end{tabular}
\end{table}

\subsection{Performance on Real-World Data}
\label{subsec:PerformanceRealData}
Examining the performance of YOLO-FEDER FusionNet on real-world datasets R1 and R2 (with varying cutout sizes, cf. Section~\ref{subsec:Data}) reveals promising results, particularly in contrast to the original YOLOv5l model trained and assessed on 2040$\times$1086 images~\cite{Dieter:2023}. Specifically, YOLO-FEDER FusionNet exhibits a significant decline in FDR of 77.2~\% (from 0.5 to 0.114) and 86.8~\% (from 0.5 to 0.066) for dataset R1 (cf. Table~\ref{tab:resultsR1}). Additionally, an exceptional FDR reduction exceeding 90.0~\% is observed for R2 across both image sizes, with values decreasing from 0.29 to 0.029 and 0.007 (cf. Table~\ref{tab:resultsR2}). Furthermore, there is also a distinct reduction in FNRs. 

The direct comparison between YOLOv5l -- trained on un-cropped images of S1 and evaluated on un-cropped images of R1 -- and YOLO-FEDER FusionNet reveals a variance in FNRs of 0.091 and 0.014, respectively. This disparity is further highlighted when contrasting the evaluation results of YOLO-FEDER FusionNet with those of YOLOv5l on cropped versions of R1 (cf. Table~\ref{tab:resultsR1}). A substantial discrepancy in FNRs becomes even more evident for dataset R2, where highly textured objects form a significant portion of the images' background (cf. Figure~\ref{fig:Comparison}). While YOLOv5l, evaluated on the original-sized R2 images (cf. 2040$\times$1086, Table~\ref{tab:resultsR2}), exhibits a FNR of 0.745, YOLO-FEDER FusionNet significantly reduces this rate. Specifically, when evaluated on 1080$\times$1080 image cutouts of R2, the FNR diminishes to less than half of its original magnitude. This observed trend remains consistent when contrasted with YOLOv5l SQ, trained similarly to YOLO-FEDER FusionNet on a cropped version of S1. However, owing to its consistently high FNRs and FDRs close to zero (cf. Tables~\ref{tab:resultsR1} and \ref{tab:resultsR2}), YOLOv5l SQ demonstrates a general inefficiency in the present context of drone detection. On the contrary, this highlights the performance advantages attained through the integration of YOLOv5l and FEDER within YOLO-FEDER FusionNet.

\begin{figure}[t!]
\begin{minipage}[b]{1\linewidth}
  \includegraphics[width=\textwidth, trim={0.1cm 16cm 0.1cm 0cm}, clip]{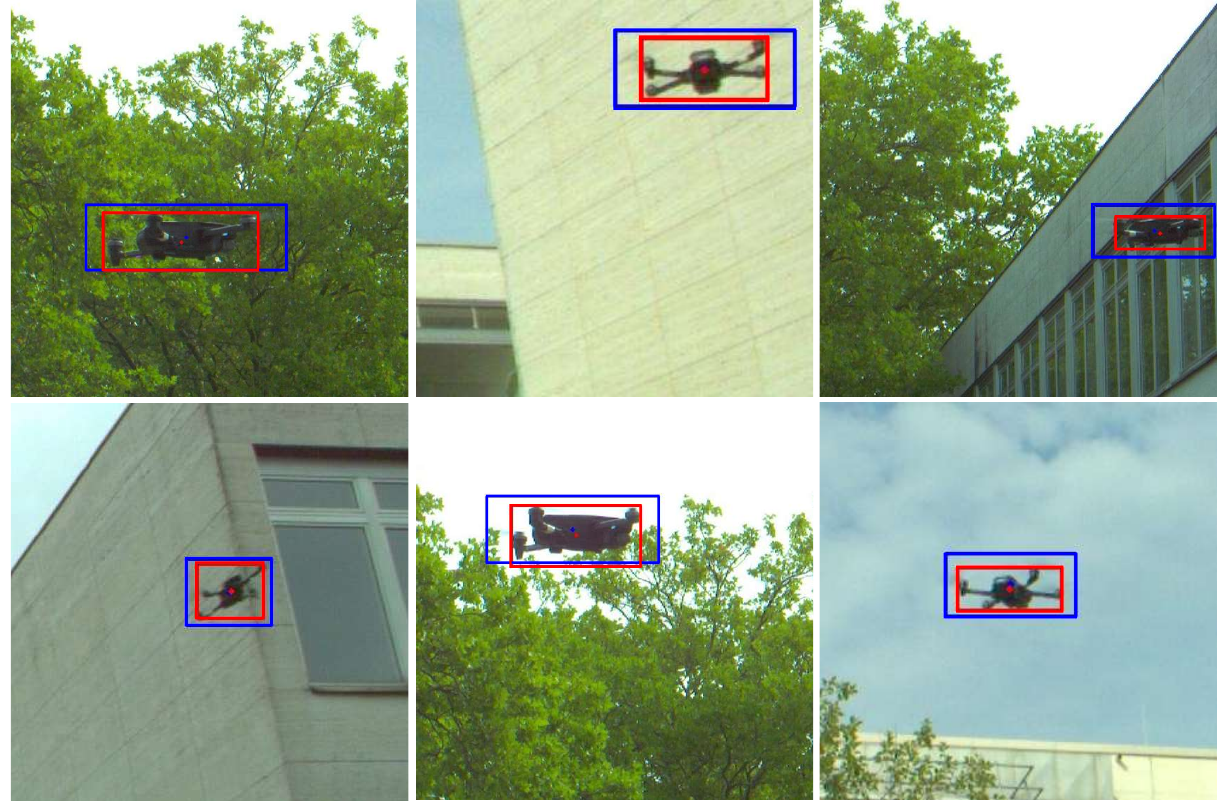}
  \caption{\label{fig:BboxAlignment} Visual comparison of the manually labeled GT bounding boxes (blue) and the bounding boxes predicted by YOLO-FEDER FusionNet (red). While the GT boxes provide a more generous encapsulation of the drone, the predicted bounding boxes demonstrate a superior level of accuracy.}
\end{minipage}
\end{figure}

Analyzing the mAP values at an IoU threshold of 0.5 reveals distinct trends. Within dataset R1, there is a significant improvement in mAP, rising from 0.559 (cf. YOLOv5l, 2040$\times$1086, Table~\ref{tab:resultsR1}) to 0.636 and 0.669 upon implementing YOLO-FEDER FusionNet. Conversely, a marginal decline in mAP values is evident within dataset R2 (cf. Table~\ref{tab:resultsR2}). Upon closer examination of the mAP values at an IoU threshold of 0.25, YOLO-FEDER FusionNet demonstrates superior performance when compared to YOLOv5l. Specifically, YOLO-FEDER FusionNet exhibits mAP values ranging between 0.685 and 0.816, while YOLOv5l solely registers values below 0.572. 

\subsection{Labeling Bias}
\label{subsec:LabelBias}
Despite the precise drone localization capabilities of YOLO-FEDER FusionNet (see Figure~\ref{fig:BboxAlignment}), a deeper analysis comparing predicted bounding boxes with the ground truth (GT) reveals a discrepancy in their spatial overlap. Unlike synthetic training data, which features pixel-precise labeling, the manual annotations of R1 and R2 seem to include more pixels than necessary to accurately localize a drone. Consequently, they tend to cover a slightly larger area to ensure comprehensive object encapsulation with high certainty. For instance, 72.71~\% of the predicted bounding boxes of R1 (640$\times$640) are entirely contained within the GT. For dataset R2 (1080$\times$1080), it's 63.45~\%. However, this bias in manual labeling significantly compromises detection quality, leading to an inferior model performance in terms of mAP (especially in scenarios involving both synthetic and real-world data). 

\begin{table}[t!]
\centering
\caption{Categorization of predicted bounding box dimensions into distinct groups, including size-dependent scaling factors for bounding box correction against labeling biases.}
  \label{tab:groups}
  \footnotesize
  \begin{tabular}{lcccccc}
  \hline\noalign{\smallskip}
    Category & \multicolumn{2}{c}{Width Ratio} & \multicolumn{2}{c}{Height Ratio} & \multicolumn{2}{c}{Scaling Factors}\\
    & min & max & min & max & $\lambda_w$ & $\lambda_h$ \\\noalign{\smallskip}\hline\noalign{\smallskip}
    Extra Small & 0.000 & 0.034 & 0.000 & 0.014 & 0.0155
& 0.0110 \\\noalign{\smallskip}\hline\noalign{\smallskip}
    Small & 0.034 & 0.059 & 0.014 & 0.027 & 0.0107 & 0.0055\\\noalign{\smallskip}\hline\noalign{\smallskip}
    Medium & 0.059 & 0.094 & 0.027 & 0.044 & 0.0071 & 0.0020 \\\noalign{\smallskip}\hline\noalign{\smallskip}
    Large & 0.094 & 0.144 & 0.044 & 0.072 & 0.0044 & 0.0014 \\\noalign{\smallskip}\hline\noalign{\smallskip}
    Extra Large & 0.144 & 1.000 & 0.072 & 1.000 & 0.0022 & 0.0011\\\hline
  \end{tabular}
\end{table}

\begin{table}[t!]
\centering
\caption{Mean average precision of YOLO-FEDER FusionNet on datasets R1 and R2 considering manual labeling biases.}
  \label{tab:labelbias}
  \footnotesize
  \begin{tabular}{lccccc}
  \hline\noalign{\smallskip}
    Dataset & Img Size & \multicolumn{2}{c}{Labeling Bias} & \multicolumn{2}{c}{mAP} \\\noalign{\smallskip}\noalign{\smallskip}
    & & Included & Type & @0.25 & @0.5 \\\noalign{\smallskip}\hline\noalign{\smallskip}
    \multirow{3}*{R1} & \multirow{3}*{640$\times$640} & \xmark & -- & 0.729 & 0.669\\
    & & \cmark & fixed & 0.729 & 0.700\\
    & & \cmark & variable & 0.729 & 0.710\\\noalign{\smallskip}\hline\noalign{\smallskip}
    
    \multirow{3}*{R1} & \multirow{3}*{1080$\times$1080} & \xmark & -- & 0.708 & 0.636\\
    & & \cmark & fixed & 0.708 & 0.666 \\
    & & \cmark & variable & 0.708 & 0.681 \\\noalign{\smallskip}\hline\noalign{\smallskip}
    
    \multirow{3}*{R2} & \multirow{3}*{640$\times$640} & \xmark & -- & 0.685 & 0.270 \\
    & & \cmark & fixed& 0.714 & 0.317 \\
    & & \cmark & variable & 0.720 & 0.416 \\\noalign{\smallskip}\hline\noalign{\smallskip}

    \multirow{3}*{R2} & \multirow{3}*{1080$\times$1080} & \xmark & -- & 0.816 & 0.423 \\
    & & \cmark & fixed & 0.827 & 0.472 \\
    & & \cmark & variable & 0.836 & 0.701\\\hline
  \end{tabular}
\end{table}

To address this issue, we propose the integration of a post-processing strategy designed to compensate for deviations stemming from manual labeling. A key advantage of this strategy lies in its capacity to obviate the necessity for modifying existing datasets and undergoing re-training. The process entails the refinement of predicted bounding boxes, characterized by width $w$ and height $h$, through a formal bias compensation approach: $w' = w + \lambda_w(w\cdot h)$ and $h' = h + \lambda_h(w\cdot h)$, where $w'$ and $h'$ signify the adjusted bounding box width and height. The scaling factors $\lambda_w$ and $\lambda_h$ can be tailored individually. In our evaluation, we considered both fixed factors ($\lambda_w=0.0057$ and $\lambda_h=0.0023$) and adaptive scaling factors linked to the object's size (cf. Table~\ref{tab:groups}). Notably, $\lambda_w$ and $\lambda_h$ diminish with larger object sizes, as smaller objects tend to exhibit more pronounced labeling bias due to the intricacy involved in their labeling process.

As illustrated in Table~\ref{tab:labelbias}, accounting for manual labeling bias enhances the mAP, especially at an IoU threshold of 0.5. A considerable improvement can be observed for dataset R2, suggesting an impact of background complexity on the extent of the labeling bias. Thus, addressing this bias seems to be particularly beneficial in scenarios characterized by intricate or highly textured backgrounds, such as trees.

\subsection{Drone Detection in an Alarm Scenario}
Drone detection can also be seen as an integral component of a comprehensive security system, specifically targeting the identification of potential drone threats and the subsequent activation of warning mechanisms. Hence, inferring the existence of a drone within a video sequence does not necessarily mandate a frame-by-frame detection. Alternatively, the presence of a drone can be inferred based on a partial sequence of frames, where its appearance in at least one frame indicates its existence. This strategy leads to a decline of missed detections (cf. Table~\ref{tab:alarmScenario}), albeit at the expense of an increased inference time.

\begin{table}[t!]
\centering
\caption{Evolution of FNR for individual camera positions of dataset R2 (1080$\times$1080) relative to the sequence size.}
  \label{tab:alarmScenario}
  \footnotesize
  \begin{tabular}{lcccccc}
  \hline\noalign{\smallskip}
    Camera Pos. & \multicolumn{6}{c}{Sequence Size (\# Frames)}\\
    & 1 & 5 & 11 & 17 & 21 & 27 \\\noalign{\smallskip}\hline\noalign{\smallskip}

    POS1 & 0.269 & 0.153 & 0.118 & 0.087 & 0.053 & 0.052\\\noalign{\smallskip}\hline\noalign{\smallskip}
    POS2 & 0.266 & 0.175 & 0.155 & 0.117 & 0.113 & 0.122\\\hline
  \end{tabular}
\end{table}

% CONCLUSION
\section{Conclusion}
\label{sec:conclusion}
In this work, we explored the effectiveness of integrating generic object detection algorithms with COD techniques for drone detection in environments with complex backgrounds. We introduced YOLO-FEDER FusionNet, a novel DL architecture. Alongside the integration of dual backbones, we implemented a redesigned neck structure to enable seamless information fusion and facilitate the prioritization of essential features. We systematically evaluated the proposed detection model on a variety of real and synthetic datasets, characterized by different complexity levels. Our analyses demonstrated substantial improvements of YOLO-FEDER FusionNet over conventional drone detectors, especially in terms of FNRs and FDRs. Furthermore, we revealed a labeling bias originating from manually generated annotations in real-world data, adversely affecting mAP values. Addressing this bias via post-processing led to improvements w.r.t. mAP. We also showed that leveraging information from previous frames in a video stream can further reduce FNRs. 

\paragraph*{Funding:} \hspace{-0.3cm} No funding was received for conducting this study.\vspace{-0.3cm}

\paragraph*{Conflicts of Interest:} \hspace{-0.3cm} The authors declare no relevant financial or non-financial conflicts of interest.

% REFERENCES
\bibliographystyle{IEEEbib}
\bibliography{refs}

\end{document}